\documentclass{article} 
\usepackage{iclr2026_conference,times}
\iclrfinalcopy

\usepackage{amsmath,amsfonts,bm}

\def\eqref#1{equation~\ref{#1}}

\def\1{\bm{1}}

\DeclareMathAlphabet{\mathsfit}{\encodingdefault}{\sfdefault}{m}{sl}
\SetMathAlphabet{\mathsfit}{bold}{\encodingdefault}{\sfdefault}{bx}{n}

\usepackage{hyperref}
\usepackage{url}
\usepackage{tikz} 
\usepackage[utf8]{inputenc} 
\usepackage[T1]{fontenc}    
\usepackage{hyperref}       
\usepackage{url}            
\usepackage{booktabs}       
\usepackage{amsfonts}       
\usepackage{nicefrac}       
\usepackage{microtype}      
\usepackage{xcolor}         
\usepackage{multirow}

\newcommand{\projname}[0]{ExoActor}

\newcommand{\ie}{\emph{i.e., }}
\newcommand{\eg}{\emph{e.g., }}

\usepackage{wrapfig}
\usepackage{graphicx}
\usepackage{amsmath}

\usepackage{times}
\usepackage{latexsym}
\usepackage{pifont}
\usepackage[table]{xcolor}   
 \usepackage{amsmath}
\definecolor{lightgray}{gray}{0.92}
\usepackage{booktabs}
\usepackage{ulem} 
\usepackage[most]{tcolorbox}
\usepackage{enumitem} 

\definecolor{boxTitleBg}{RGB}{80, 120, 160}    
\definecolor{boxContentBg}{RGB}{240, 245, 250} 
\definecolor{boxFrameColor}{RGB}{80, 120, 160} 

\newtcolorbox{systemprompt}[1][]{
  enhanced,               
  title={System Instruction}, 
  colframe=boxFrameColor, 
  colbacktitle=boxTitleBg,
  coltitle=white,         
  colback=boxContentBg,   
  fonttitle=\bfseries\rmfamily, 
  fontupper=\rmfamily,    
  arc=1.5mm,              
  boxrule=0.7pt,          
  titlerule=0pt,          
  left=5pt, right=5pt, top=5pt, bottom=5pt, 
  #1 
}

\title{\projname{}: Exocentric Video Generation as Generalizable Interactive Humanoid Control}

\author{%
  Yanghao Zhou$^*$, Jingyu Ma\thanks{Equal Contribution. Work done during their internships at BAAI.}  , Yibo Peng, Zhenguo Sun,
  \textbf{Yu Bai\thanks{Corresponding Author: \url{baiyu@baai.ac.cn}.} , Börje F. Karlsson}  \\ 
    Beijing Academy of Artificial Intelligence (BAAI)\\ 
}

\begin{document}

\maketitle

\begin{tikzpicture}[remember picture, overlay]
    \node[anchor=north west, yshift=-0.3in, xshift=1.0in] at (current page.north west) {
        \includegraphics[width=3cm]{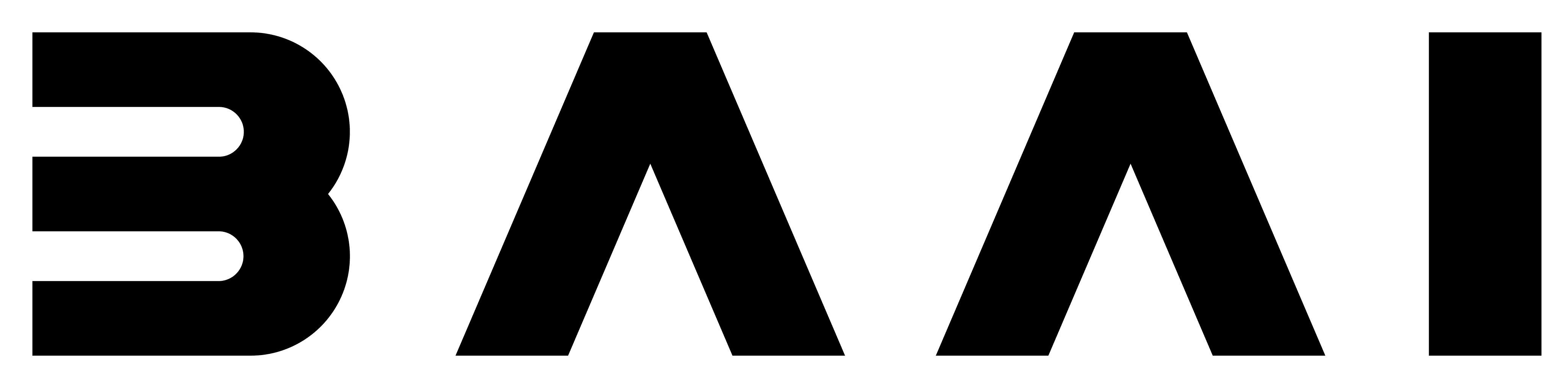}
    };
    
    \draw[darkgray] (current page.north west) ++(0.95in, -0.7in) -- ++(\columnwidth * 1.2, 0);
\end{tikzpicture}
\begin{abstract}

Humanoid control systems have made significant progress in recent years, yet modeling fluent interaction-rich behavior between a robot, its surrounding environment, and task-relevant objects remains a fundamental challenge. This difficulty arises from the need to jointly capture spatial context, temporal dynamics, robot actions, and task intent at scale, which is a poor match to conventional supervision. We propose \projname{}, a novel framework that leverages the generalization capabilities of large-scale video generation models to address this problem. The key insight in \projname{} is to use third-person video generation as a unified interface for modeling interaction dynamics. Given a task instruction and scene context, \projname{} synthesizes plausible execution processes that implicitly encode coordinated interactions between robot, environment, and objects. Such video output is then transformed into executable humanoid behaviors through a pipeline that estimates human motion and executes it via a general motion controller, yielding a task-conditioned behavior sequence. 
To validate the proposed framework, we implement it as an end-to-end system and demonstrate its generalization to new scenarios without additional real-world data collection.
Furthermore, we conclude by discussing limitations of the current implementation and outlining promising directions for future research, illustrating how  \projname{} provides a scalable approach to modeling interaction-rich humanoid behaviors, potentially opening a new avenue for generative models to advance general-purpose humanoid intelligence\footnote{Project page:~\url{https://baai-agents.github.io/ExoActor/}.}.

\end{abstract}

\section{Introduction}

Humanoid robots hold the promise of operating in unstructured human environments and performing a wide range of tasks with minimal task-specific engineering or supervision~\citep{bai2026egoactor0, yuan2025being000}. Despite recent advances in learning-based control and whole-body coordination~\citep{luo2025sonicsupersizingmotiontracking, liao2025beyondmimic0}, enabling fluent interactions between the robot, its surrounding environment, and task-relevant objects remains a central challenge. Such interaction-rich behaviors require jointly modeling spatial context, temporal dynamics, and task intent, which current systems struggle to generalize across diverse environments and tasks~\citep{ma2026generalvla0, he2025viral0}. As a result, policies trained in controlled settings often fail in new scenes, and scaling to broader task distributions typically relies on expensive data collection, domain-specific tuning, or carefully curated demonstrations.

\begin{figure*}
\centering
\includegraphics[width=0.9\textwidth]{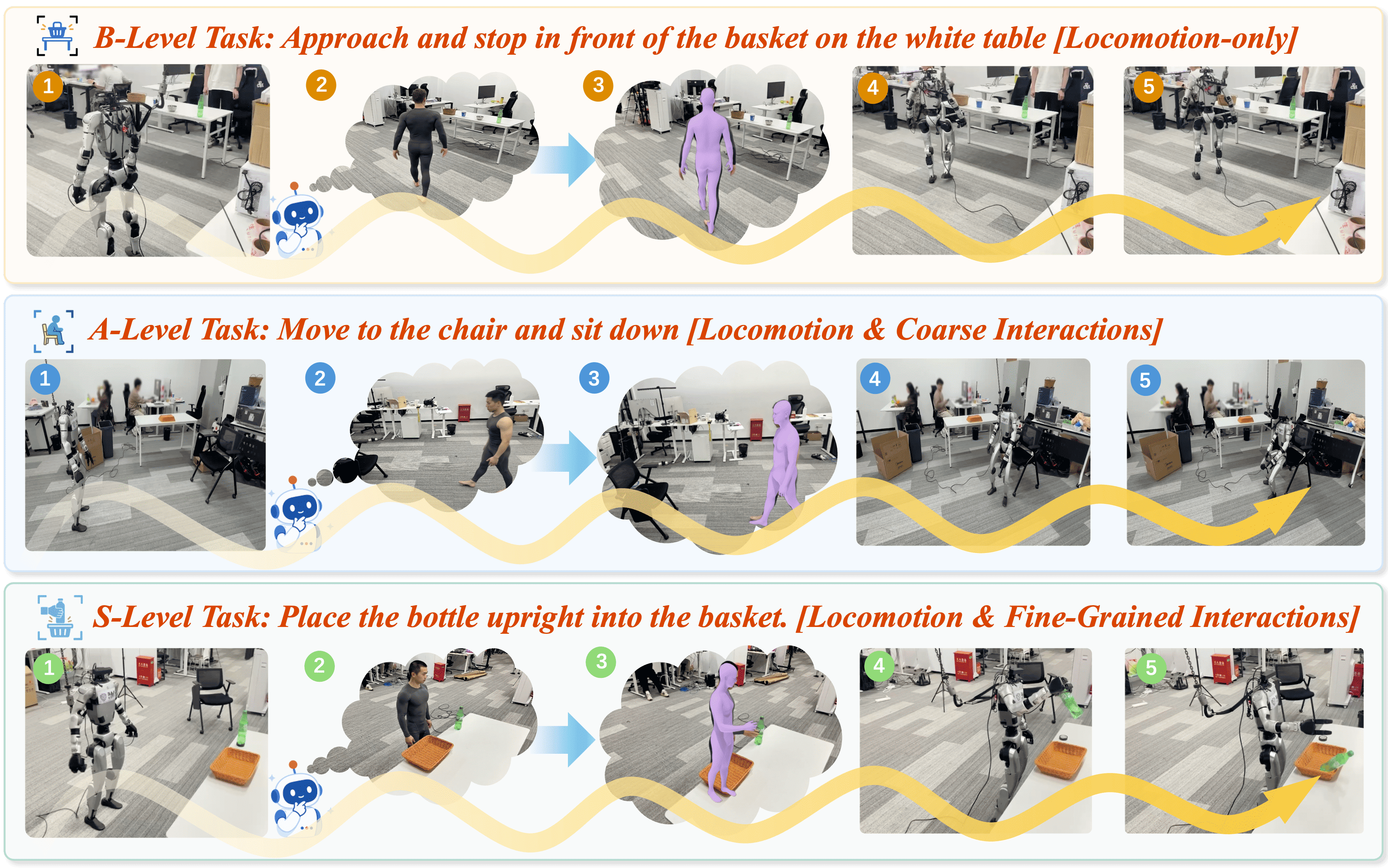} 
\caption{\projname{} capabilities across tasks of increasing difficulty. Exocentric video generation is used to produce imagined demonstrations, which are then converted into executable humanoid behaviors. Examples span different difficulty levels (\ie \textbf{B-level} (easy), \textbf{A-level} (moderate), and \textbf{S-level} (challenging), see Section~\ref{sec:definition}), highlighting its generalization in diverse interaction-rich tasks.}
\vspace{-0.5em}
\label{fig:intro}
\end{figure*}

In this work, we propose \projname{}, a framework and system that bridges generative video modeling and humanoid control through third-person video generation~\citep{ye2026world, bruce2024genie0, chen2025large}, which has been shown to exhibit strong generalization capabilities~\citep{team2025kling0omni, seedance2026seedance, videoworldsimulators2024}. As illustrated in Fig.~\ref{fig:intro}, given a task description and an initial third-person observation, our framework synthesizes a plausible execution process in the form of third-person video plans that captures coordinated interactions between the robot, environment, and target objects. These videos serve as high-level ``imagined demonstrations,’’ which are then transformed into executable robot behaviors through a unified pipeline consisting of whole-body and hand motion estimation, followed by end-to-end motion execution. By decoupling high-level interaction modeling from low-level control, \projname{} enables humanoid systems to leverage the generalization capabilities of video models while remaining compatible with existing control frameworks.

The \projname{} approach offers several advantages. First, it removes the need for task-specific data collection by exploiting the implicit knowledge embedded in pretrained video models. Second, third-person video provides a flexible and expressive interface that naturally captures long-horizon behaviors and rich interaction context. Third, the modular pipeline—video generation, motion estimation, and execution—allows each component to be improved independently, making the system extensible and adaptable. As shown in Fig.~\ref{fig:intro}, this design enables a unified transition from visual imagination to physical execution across tasks of increasing complexity.

We implement \projname{} as an end-to-end system and demonstrate its feasibility in enabling humanoid robots to execute a diverse set of tasks derived purely from generated videos. While our results highlight the potential of this paradigm, they also expose several important challenges and directions for future work. We believe that \projname{} represents a step toward a new class of humanoid embodied systems that leverage generative models not only for perception or planning, but also as a scalable interface for synthesizing novel behavior (including the generation of counterfactuals). By connecting high-level generative priors with low-level control execution, the proposed framework suggests a promising pathway toward more generalizable humanoid intelligence.

We summarize our contributions as follows:
\begin{itemize}
    \item We identify exocentric (third-person) video generation as a scalable paradigm for modeling interaction dynamics in humanoid control, effectively leveraging the generalization capabilities of large pretrained video models.
    \item We propose \textbf{\projname{}}, an end-to-end framework that synthesizes task execution videos and directly converts them into executable humanoid behaviors via human motion estimation and general motion tracking, without task-specific data collection.
    \item Our implemented system demonstrates the feasibility of this paradigm on real-world humanoid tasks, showing that generated videos can be translated into interaction-aware behaviors across diverse scenarios and levels of difficulty.
    \item We discuss key limitations, challenges, and future directions, including physically grounded video generation, improved motion execution pipelines, integration with vision-based whole-body control, and extensions to manipulation-intensive tasks.
\end{itemize}

\section{Method}
\subsection{Overall Design}

\begin{figure*}
\centering
\includegraphics[width=\textwidth]{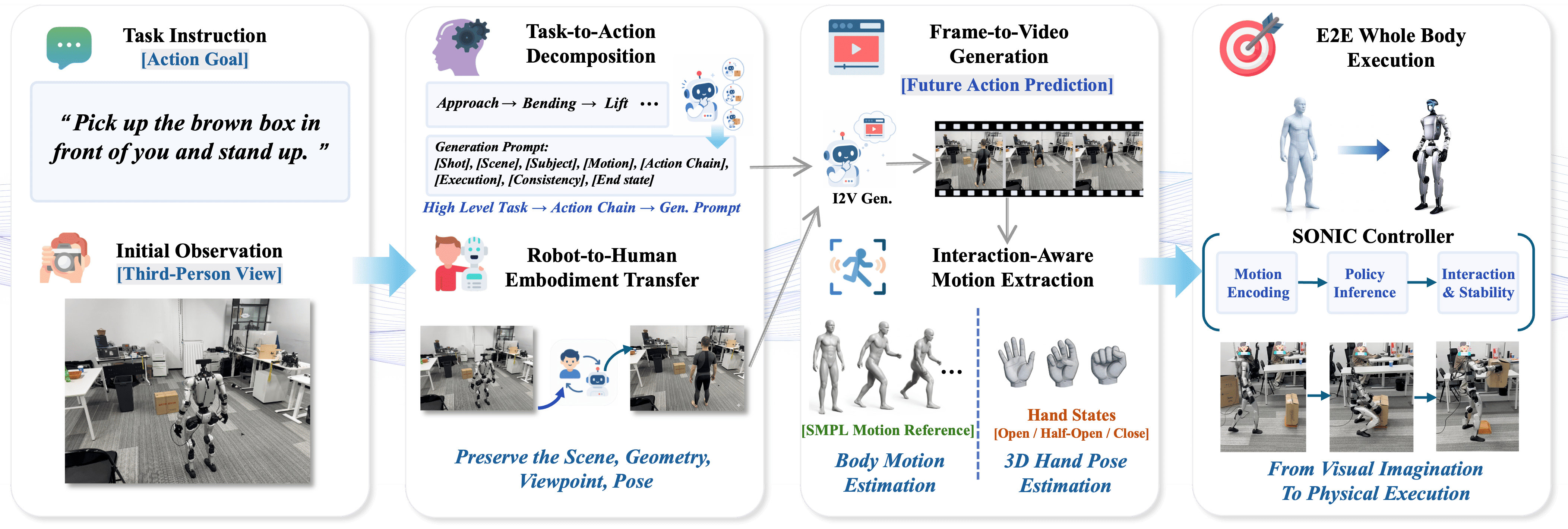} 
\caption{Overview of the \projname{} framework. Given a task instruction and initial observation, the system generates third-person action videos via embodiment transfer and action decomposition, extracts structured motion, and executes it with a motion tracking controller, enabling a transition from visual imagination to physical execution.}
\vspace{-0.5em}
\label{fig:main}
\end{figure*}

We propose \textbf{\projname{}}, a unified framework and system that converts high-level task instructions into executable humanoid behaviors by leveraging third-person video generation as intermediate representation. The overall pipeline consists of three key stages:
\textbf{1)}~\emph{Video generation:} Given a task and an initial third-person observation, the system first generates task-consistent videos depicting the execution process. To align with the human-centric priors of video generation models, we introduce a robot-to-human embodiment transfer step that transforms the robot observation into a human-like representation while preserving scene geometry and viewpoint. Conditioned on this reference and step-wise decomposed action prompts, the system produces temporally structured video clips.
\textbf{2)}~\emph{Motion estimation:} The generated videos are then converted into structured motion representations via motion estimation, recovering 3D human kinematics, without robot-specific retargeting.
\textbf{3)}~\emph{Motion execution:} A general motion executor directly consumes the estimated motions to produce dynamically feasible and stable robot behaviors. An overview is illustrated in Fig.~\ref{fig:main}.

\subsection{Third-person Video-Action Generation}

Our goal is to generate third-person, fixed-camera action videos that are both task-consistent and compatible with the embodiment constraints of the Unitree G1 robot. Directly prompting a video generation model to produce robot actions, however, often leads to unstable motion, distorted body geometry, and poor temporal consistency, especially when the target embodiment differs substantially from the human-centric visual priors of current video generators. To address this issue, we formulate third-person video-action generation as a two-stage process: \emph{robot-to-human embodiment transfer} followed by \emph{task- and environment-generalizable video generation}. The overall pipeline first transforms a robot-centered visual reference into a human-compatible initialization, then decomposes a high-level task instruction into executable step-wise action prompts, and finally generates a sequence of third-person action videos under a fixed camera view.

\paragraph{Robot-to-Human Embodiment Transfer}

A key challenge in our setting arises from the embodiment mismatch between humanoid robots and the human-centric priors of large-scale video generation and motion estimation models. Since these models are predominantly trained on human data, directly conditioning on robot appearances often leads to unstable video generation with hallucinations and degraded motion estimation accuracy.

\begin{wrapfigure}{r}{0.53\textwidth}
\vspace{-2em}
\centering
\includegraphics[width=\linewidth]{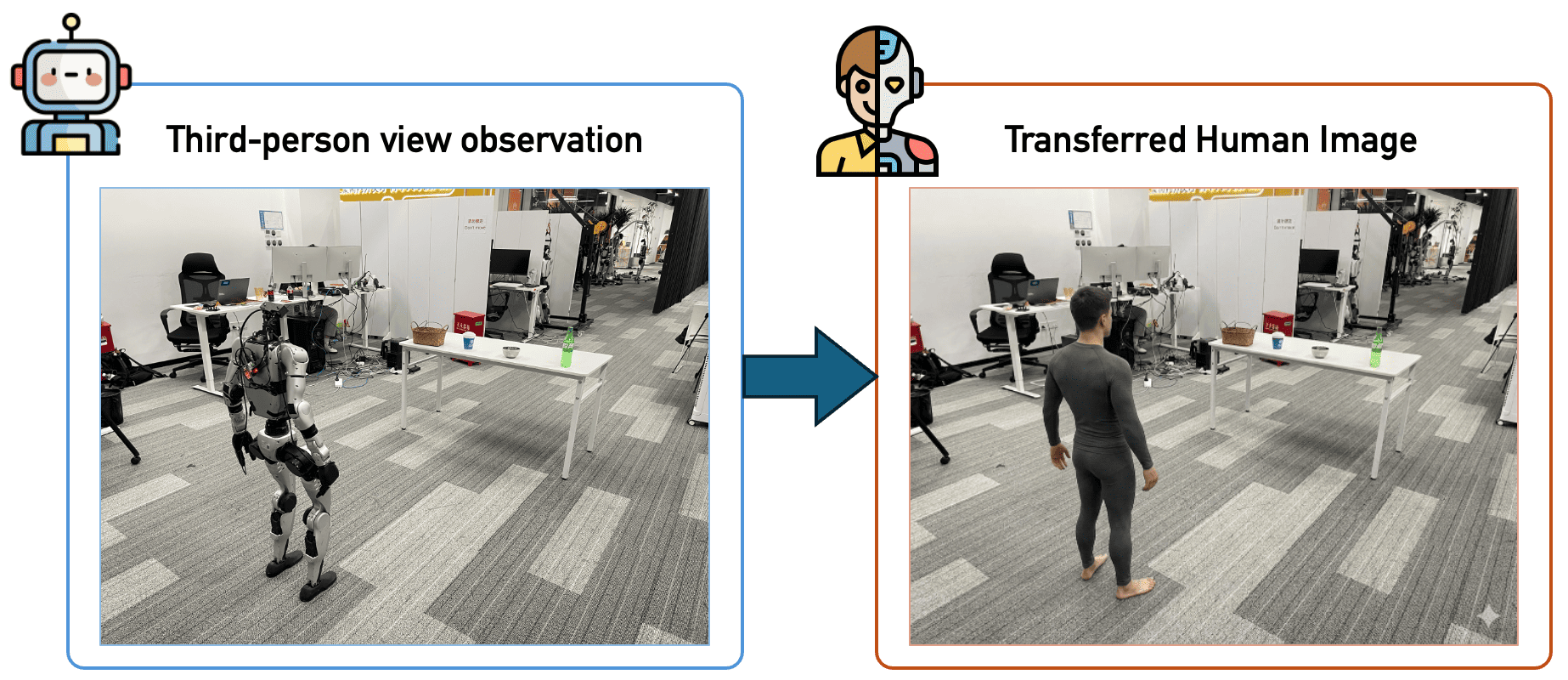} 
\caption{Illustration of robot-to-human embodiment transfer. We convert a robot third-person view observation (left) into a human-like representation (right) while preserving scene layout, viewpoint, and pose, enabling more stable video generation.}
\vspace{-0.5em}
\label{fig:style_transfer}
\end{wrapfigure}

To address this issue, we introduce a \emph{robot-to-human embodiment transfer} step (Fig.~\ref{fig:style_transfer}), which converts the robot-centric input into a human-like representation that aligns with these learned priors. Starting from a third-person image of the robot in the target scene, we transform the robot into a human subject while strictly preserving the original scene layout, camera viewpoint, body pose, orientation, scale, and robot-body proportions.
This transformation maps the input into a domain where both video generation and motion estimation models operate more reliably. Empirically, it improves temporal consistency, reduces visual artifacts, and produces more physically plausible video sequences. It also enhances motion estimation accuracy by aligning inputs with human-centric training distributions, thereby improving the overall stability of the pipeline.

In our implementation, this transformation is performed using Nano Banana Pro (Gemini 3.1 Pro) via a prompt-based image editing interface. The prompt enforces strict constraints on pose, orientation, and scene preservation, and is provided in Appendix~\ref{app:robot_to_human_prompt}.

\paragraph{Task-to-Action Decomposition and Prompt Construction.}
A key challenge in video-driven humanoid control is that high-level task instructions are often abstract and underspecified, making them difficult to directly translate into controllable and temporally coherent video generation. To address this, we decompose each instruction into an explicit sequence of intermediate actions, providing structured guidance for downstream generation.

Specifically, we convert each high-level instruction $G$ into a temporally ordered action chain $C=\{a_1,a_2,\ldots,a_T\}$, where each $a_t$ represents an atomic, visually observable, and physically executable action. This decomposition preserves task intent, object attributes, and spatial relationships while removing redundant or ambiguous steps. For example, ``Pick up the brown box in front of you and stand up'' is decomposed into \emph{approach the box} $\rightarrow$ \emph{bend down} $\rightarrow$ \emph{grasp the box} $\rightarrow$ \emph{lift the box} $\rightarrow$ \emph{stand upright}. 
We then construct an observation-aware generation prompt by grounding the action chain in the initial observation and task goal. Specifically, the action chain provides the temporal structure of the behavior, while the observation supplies scene layout, object locations, and spatial constraints. These inputs are fused into a scene- and task-aware action description that explicitly aligns each sub-action with the visible environment.

In our implementation, both action decomposition and prompt construction are performed using GPT-5.4 Thinking. The full prompt design and decomposition format are provided in Appendix~\ref{app:action_decomposition} and the action description construction prompt is provided in Appendix~\ref{app:fusion_prompt}

\paragraph{Task- and Environment-Generalizable Video Generation}

Given the transferred human-like reference frame, we generate task-consistent third-person videos using a structured prompt template. A key challenge at this stage is that video generation models often produce hallucinated content or inconsistent behaviors when conditioned on vague inputs. 

To address this, we design prompt templates that explicitly encode scene constraints, motion requirements, and task execution details. The templates take the initial observation and enriched action description as inputs, and organize them into structured fields such as \emph{Shot}, \emph{Scene}, \emph{Motion}, \emph{Execution}, and \emph{End State}. These fields act as explicit constraints that enforce fixed camera viewpoints, preserve scene geometry, and encourage natural, physically plausible, and robot-aligned motion patterns, thereby reducing hallucinations and improving temporal consistency.

In our implementation, we use off-the-shelf video generation APIs and primarily adopt Kling as the final model due to its superior stability and consistency; a comparison with other models is provided in Section~\ref{sec:ablation}. We use different templates for tasks of varying difficulty levels, and the prompt templates for this stage are described in Appendix~\ref{app:b_level_navigation_prompt} and~\ref{app:as_level_task_prompt}.

\subsection{Interaction-aware Motion Estimation}

To bridge the gap between pixel-level video synthesis and structured robot control, we implement an interaction-aware whole-body motion estimation module. The primary objective of this stage is to recover the 3D kinematic trajectories of the human actor from the generated third-person videos. These videos inherently encode interaction dynamics with the surrounding environment—such as object contacts, spatial constraints, and task-relevant movements—allowing the estimated motions to capture not only body kinematics but also interaction-aware behaviors.

\begin{wrapfigure}{r}{0.53\textwidth}
\vspace{-1em}
\centering
\includegraphics[width=\linewidth]{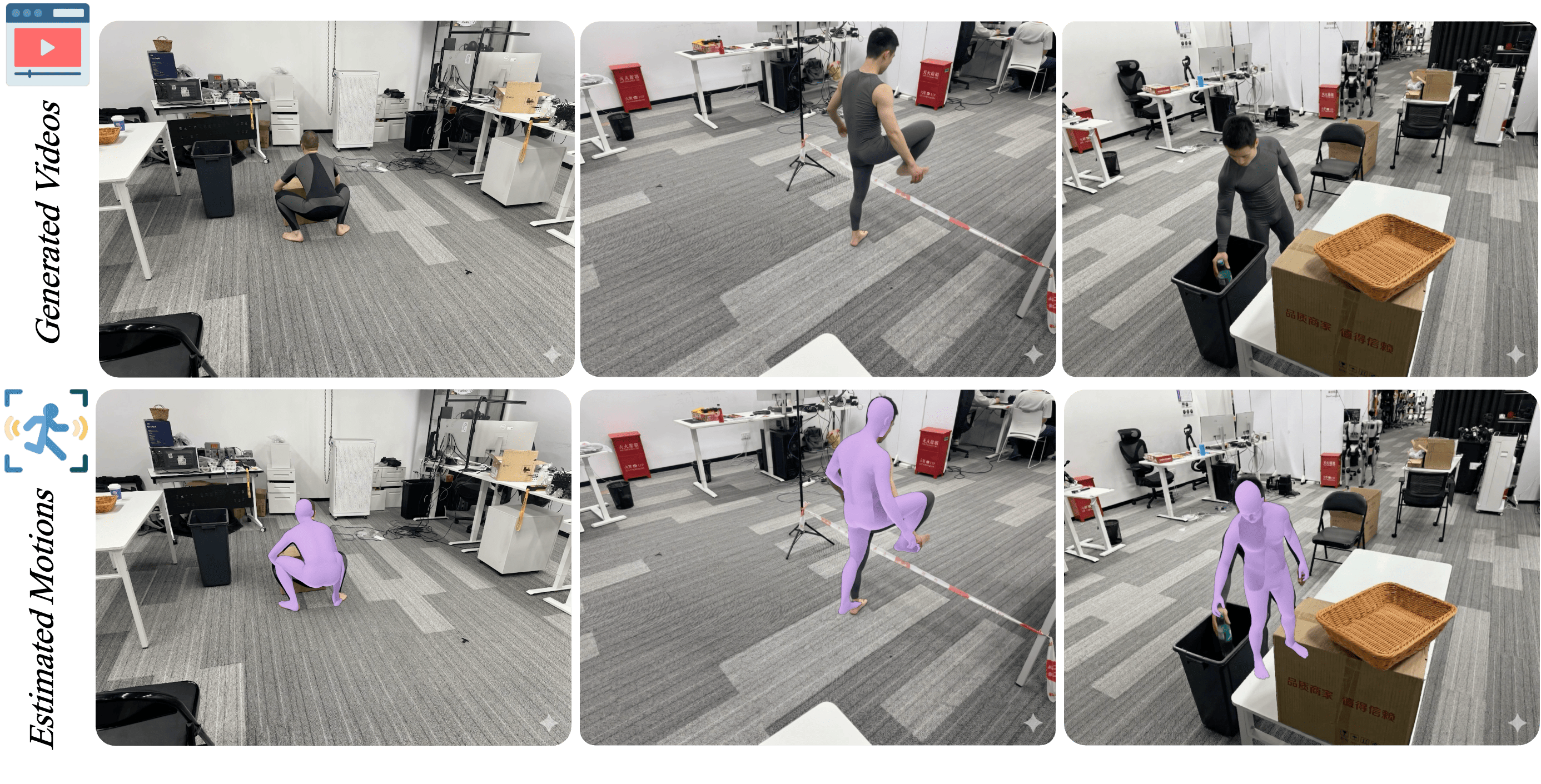} 
\caption{Interaction-aware whole-body motion estimation. Given generated third-person videos (top), our method recovers corresponding 3D human motion trajectories (bottom), capturing both body kinematics and interaction dynamics with the environment as modeled in the generated videos.}
\vspace{-0.5em}
\label{fig:whole_body_motion_extraction}
\end{wrapfigure}

\paragraph{Whole-body Motion Estimation} 
To obtain high-fidelity human motion from monocular video, we employ GENMO~\citep{li2025genmo0}, a diffusion-based model that formulates motion estimation as a constrained generation process. Instead of direct pose regression, the model leverages video features and 2D keypoints as conditioning signals to produce temporally consistent and physically plausible 3D motion sequences. In our pipeline, we use its estimation mode to directly infer motion trajectories from the generated third-person videos.

We represent the estimated motion using SMPL parameters~\citep{SMPL:2015}, including joint rotations and global positions. The model also performs temporal in-filling for partially occluded frames, producing a smooth and physically consistent motion sequence $\mathcal{M} = \{q_t, p_t\}_{t=1}^T$. This interaction-aware motion serves as a structured representation of the generated videos and provides a reliable interface for downstream robot execution commands.

\paragraph{Hand Motion Estimation}
\begin{wrapfigure}{r}{0.25\textwidth}
\vspace{-2em}
\centering
\includegraphics[width=\linewidth]{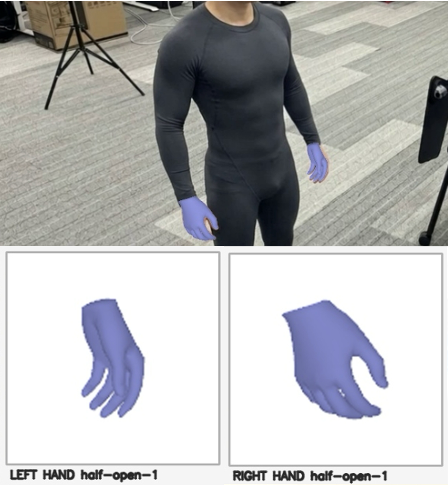} 
\caption{Hand motion estimation from generated third-person videos. Given the synthesized human observation (top), our method recovers fine-grained 3D hand poses (bottom).}
\vspace{-0.5em}
\label{fig:hand_motion_estimation}
\end{wrapfigure}

While the estimated whole-body motion $\mathcal{M}=\{q_t,p_t\}_{t=1}^{T}$ captures global body kinematics and limb coordination, it does not explicitly encode dexterous hand motion required for manipulation, such as grasp formation, release, and contact transitions. We therefore apply WiLoR~\citep{potamias2025wilor} frame by frame to the generated third-person video to estimate bilateral hand poses. We denote the tracked hand pose sequence as $\mathcal{H}=\{\mathbf{h}_t^l,\mathbf{h}_t^r\}_{t=1}^{T}$, where $\mathbf{h}_t^l$ and $\mathbf{h}_t^r$ represent the left- and right-hand pose descriptors at frame $t$, respectively. The hand annotations are stored as a $[T,2]$ array. For front-facing subjects, the two columns correspond to the anatomical left and right hands; for back-facing subjects, they correspond to the image-plane left and right hands. This viewpoint-conditioned convention mitigates handedness ambiguity and maintains a consistent correspondence between observed human hands and robot end-effectors.

Hand poses are estimated at the original video frame rate, yielding one prediction per video frame; for example, a $24$ FPS video produces a $24$ Hz hand pose sequence. Each hand is further assigned a discrete interaction state, yielding $\mathcal{S}=\{s_t^l,s_t^r\}_{t=1}^{T}$, where $s_t^l,s_t^r\in\{0,1,2\}$ correspond to \textit{open}, \textit{half-open}, and \textit{closed}~\footnote{Finer-grained hand/finger states are being addressed in future work.}. Both smoothed hand poses and interaction states are synchronized with the whole-body motion and mapped frame-wise to robot end-effector commands. Finally, we obtain an interaction-aware motion representation $\tilde{\mathcal{M}}=\{q_t,p_t,\mathbf{h}_t^l,\mathbf{h}_t^r,s_t^l,s_t^r\}_{t=1}^{T}$, which jointly encodes global body motion, bilateral hand poses, and discrete manipulation states for downstream humanoid motion execution.

\subsection{General Motion Tracking Deployment}

The final stage of the \projname{} pipeline is to transform the estimated motions into physically consistent and dynamically feasible control policies. Although the extracted motions provide high-quality kinematic references, they lack dynamical awareness—such as contact forces, torque limits, and balance constraints—required for stable execution in a physics simulator or on a physical humanoid platform.

To address this, we leverage SONIC~\citep{luo2025sonicsupersizingmotiontracking} as our motion tracking controller.
The tracking policy $\pi$ takes the current robot state $\mathbf{s}_t$ and a temporal window of reference motions $\hat{\mathbf{q}}_{t:t+k}$ (derived from our video-to-robot pipeline) as input. By utilizing SONIC’s scaled-up architecture—which significantly improves tracking precision and robustness compared to smaller-scale controllers—our system effectively "physics-filters" the "imagined" demonstrations from the video generation model. This foundation model approach ensures that even for complex or unseen parkour-style movements generated by a video generation model, our target Unitree G1 robot can maintain dynamic stability and stylistic fidelity without requiring task-specific reward engineering. The resulting controller thus achieving a seamless transition from visual "imagination" to physical "execution".
The current system implementation does not require retargeting (see Section~\ref{sec:ablation} for details).

For hand motion deployment, we map the estimated hand states from video to Unitree's Dex3-1\footnote{\url{https://www.unitree.com/Dex3-1}}compatible 7-DoF joint targets, and stream them together with the SMPL body trajectory through a event queue input interface to the robot. On the robot side, a controller decodes left and right hand joints and publishes them to specific hands at runtime. 

\begin{figure*}
\centering
\includegraphics[width=\textwidth]{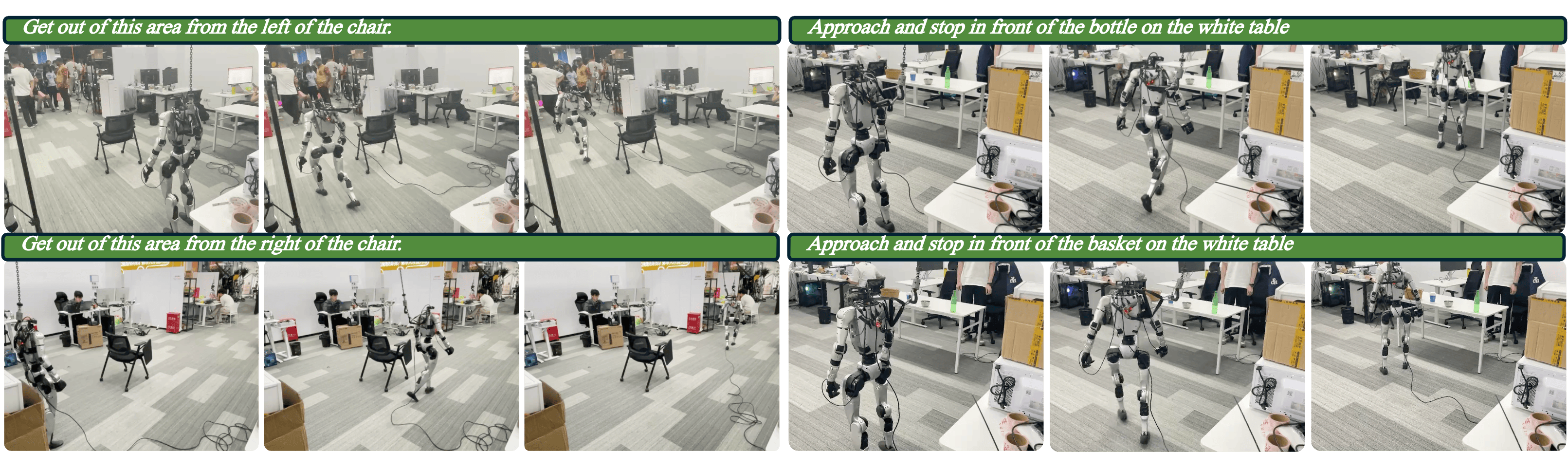} 
\caption{Case study of tasks at the \textbf{B (Easy)} difficulty level. These tasks focus on basic navigation and simple interactions in the environment.}
\vspace{-0.5em}
\label{fig:b_level}
\end{figure*}

\section{Experiments}

We evaluate \projname{} through a series of real-world experiments designed to assess its effectiveness in generating and executing interaction-rich humanoid behaviors. We first present a diverse set of tasks to demonstrate the feasibility of the proposed framework in real-world deployment. We also analyze representative success cases and conduct ablation studies to understand the impact of key design choices across different components.

\subsection{Task Definitions}
\label{sec:definition}

To evaluate the generalization capability of \projname{}, we design a set of zero-shot tasks with increasing difficulty, categorized into three levels based on the required complexity of navigation and interaction: \textbf{Level B (Easy)}, \textbf{Level A (Moderate)}, and \textbf{Level S (Challenging)}.

\paragraph{Level B (Easy).}
These tasks focus on basic locomotion and goal-directed navigation with minimal interaction. The robot is required to reach target locations or avoid simple obstacles, such as navigating toward objects (\eg a bottle or basket on a table) and moving through the environment while avoiding or passing obstacles using only navigation movements (\eg walking around chairs).

\paragraph{Level A (Moderate).}
These tasks involve navigation combined with coarse interactions with objects or the environment, without requiring precise manipulation. Typical examples include approaching objects and performing simple interactions (\eg sweeping a bottle aside), coordinating navigation with body actions (\eg moving to a chair and sitting down, lifting a box and standing up), and navigating through obstacles with whole-body movements (\eg walking under or over barriers).

\paragraph{Level S (Challenging).}
These tasks require fine-grained, multi-step interactions and more precise coordination between locomotion and manipulation. Examples include multi-step object manipulation (\eg picking up objects and placing them into a basket or trash bin) and tasks requiring accurate hand-object coordination and spatial precision (\eg placing a bottle upright into a basket).

This task hierarchy reflects increasing demands in spatial reasoning, interaction complexity, and execution precision, enabling a structured evaluation of video-driven humanoid control.

\begin{figure*}
\centering
\includegraphics[width=\textwidth]{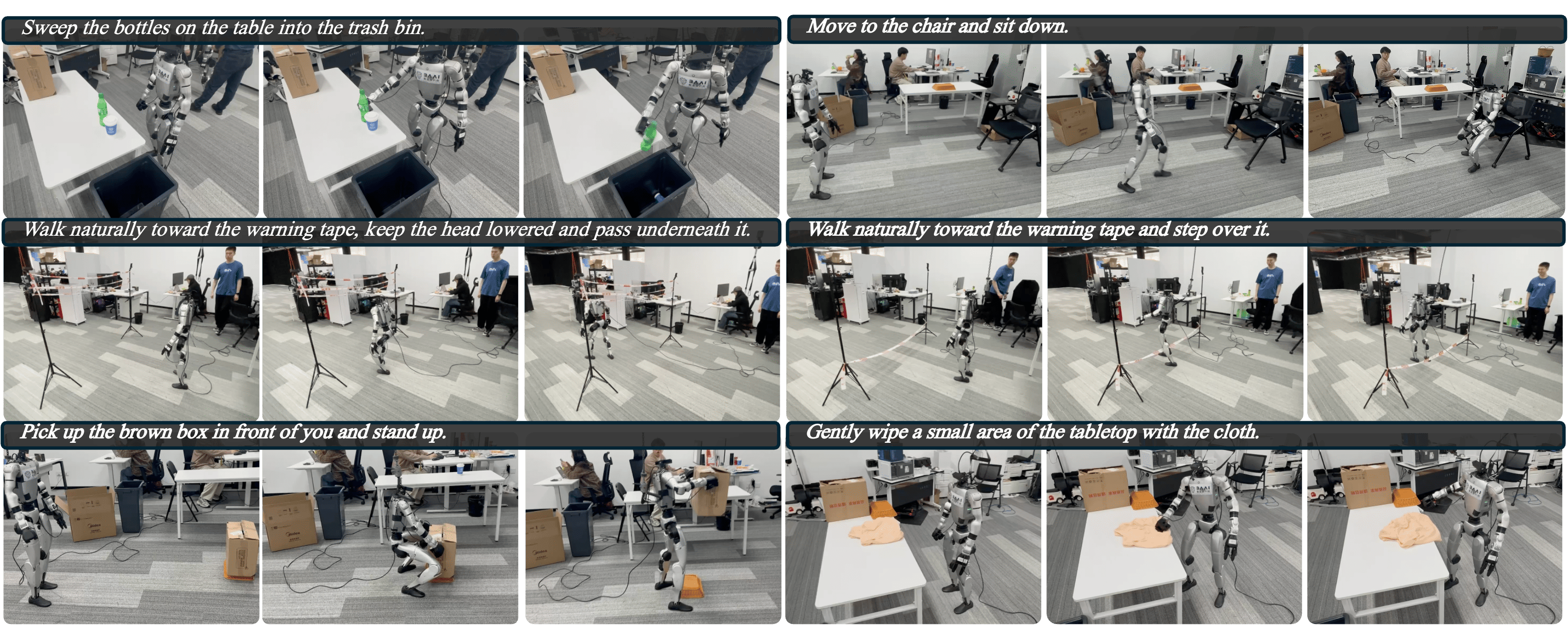} 
\caption{Case study of tasks at the \textbf{A (Moderate)} difficulty level. These tasks require the robot to perform coarse interactions with the environment and target objects.}
\vspace{-0.5em}
\label{fig:a_level}
\end{figure*}
\subsection{Case Studies}
\subsubsection{Difficulty-level Case Studies}
\label{sec:difficult_level}

\paragraph{B-tier case study.}
At the \textit{B (Easy)} level, the robot performs basic navigation tasks with minimal interaction. As shown in the Fig.~\ref{fig:b_level}, the robot can reliably approach target objects (\eg a bottle or basket on a table) and navigate around obstacles such as chairs. These tasks primarily evaluate the system’s ability to translate generated motion into stable locomotion and spatially consistent goal-reaching behavior.

\paragraph{A-tier case study.}
At the \textit{A (Moderate)} level, tasks involve coordinated navigation and coarse interactions with the environment. The robot demonstrates the ability to perform actions such as sweeping items into a trash bin, sitting on a chair, passing under or over obstacles, lifting a box, and wiping a tabletop. These examples shown in Fig.~\ref{fig:a_level} highlight that the system can execute multi-step behaviors that combine locomotion with whole-body interactions, without requiring precise dexterous manipulation.

\begin{figure*}
\centering
\includegraphics[width=\textwidth]{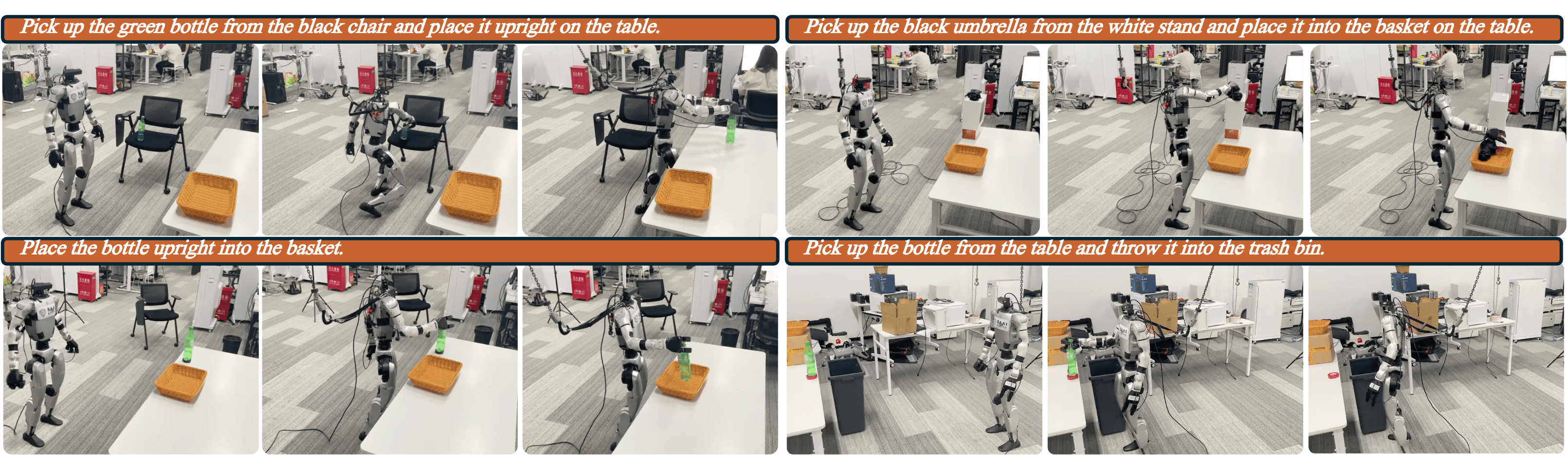} 
\caption{Case study of tasks at the \textbf{S (Challenging)} difficulty level. These tasks involve fine-grained interactions with the environment and objects.}
\vspace{-0.5em}
\label{fig:s_level}
\end{figure*}

\paragraph{S-tier case study.}
At the \textit{S (Challenging)} level, tasks require fine-grained and multi-step manipulation. The robot is able to pick up objects at different heights and place them into target containers, such as placing a bottle upright into a basket or throwing it into a trash bin, shown in Fig.~\ref{fig:s_level}. These tasks demand accurate coordination between locomotion and hand-object interaction. In practice, \textbf{due to residual inaccuracies in motion estimation—particularly in hand height—it is sometimes necessary to place small supporting bases under target items to compensate for height discrepancies during execution} to ensure action success. We further discuss this current limitation of the implemented system in Section~\ref{sec:discussion}.

\subsubsection{Failure Case Studies}

\begin{wrapfigure}{r}{0.53\textwidth}
\vspace{-1em}
\centering
\includegraphics[width=\linewidth]{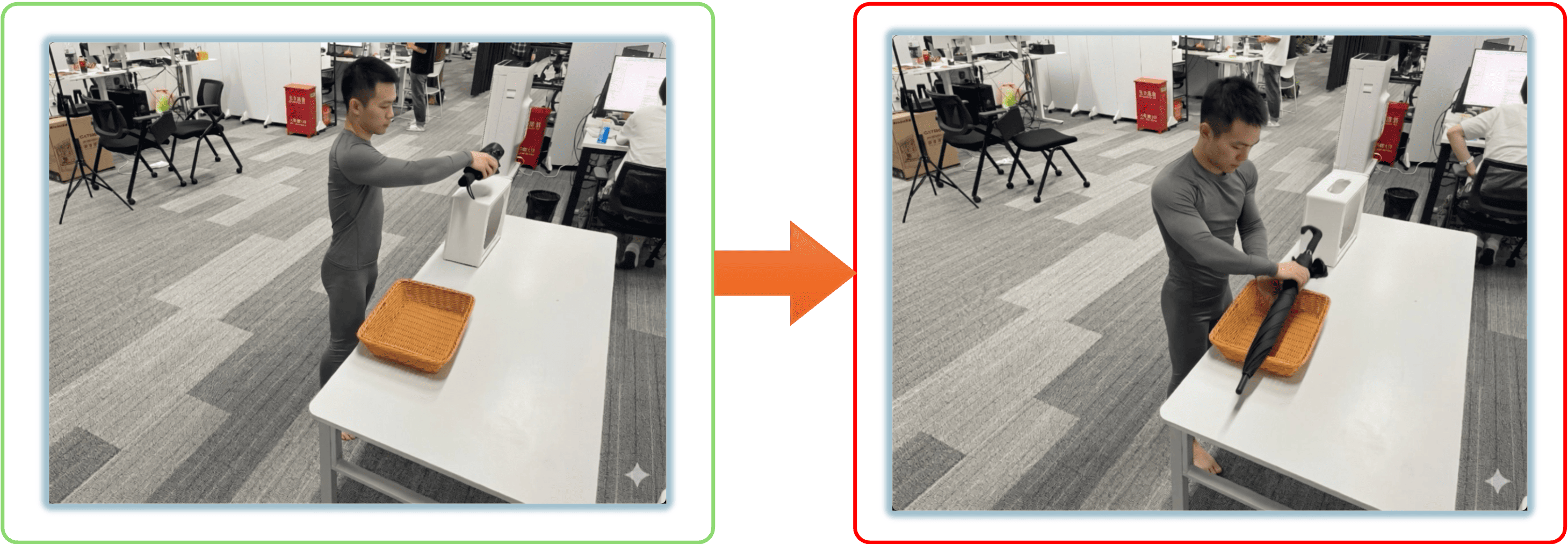} 
\caption{Failure case of video generation: a small umbrella is hallucinated into a much larger object, leading to incorrect interaction dynamics.}
\vspace{-0.5em}
\label{fig:umbrella_case}
\end{wrapfigure}

\paragraph{Failure of Video Generation}
We observe several typical failure modes in the video generation stage, including hallucination of extra objects, incorrect or inconsistent action sequences, unrealistic environment configurations, and physically implausible ending postures. For example, as shown in Fig.~\ref{fig:umbrella_case}, a small umbrella may be hallucinated as a much larger one by the video generation model. These issues can propagate to downstream stages, leading to errors in motion estimation and execution. In practice, we find that careful prompt engineering and step-wise action decomposition significantly reduce the frequency of such failures, resulting in more stable and task-consistent video outputs.

\paragraph{Failure of Motion Estimation}

\begin{wrapfigure}{r}{0.35\textwidth}
\vspace{-1em}
\centering
\includegraphics[width=\linewidth]{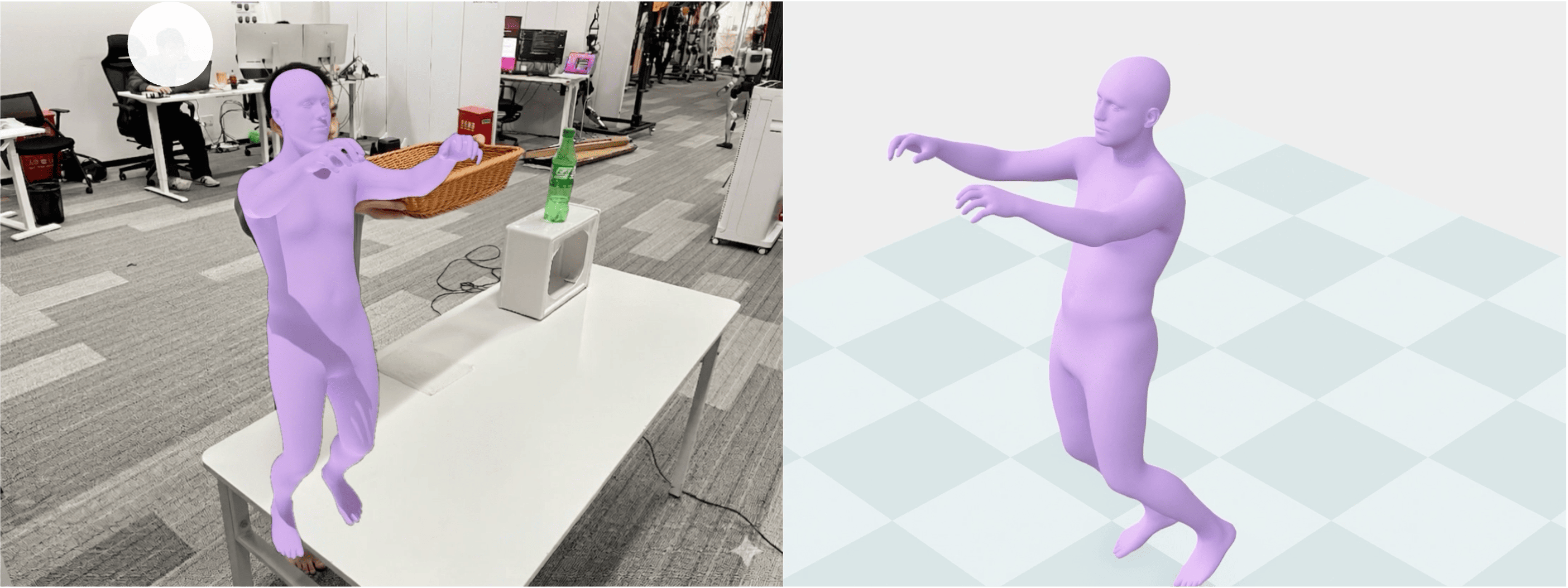} 
\caption{Failure case of motion estimation. When parts of the body are obstructed by scene objects (left), the recovered whole-body motion becomes inaccurate and incomplete (right).}
\vspace{-0.5em}
\label{fig:motion_failure}
\end{wrapfigure}

The motion estimation stage may suffer from inaccuracies in fine-grained details, particularly in hand orientation and partially-occluded body parts. For example, incorrect estimation of hand direction or ambiguous poses under occlusion can introduce errors in the recovered motion trajectories. These issues are more pronounced in fast movements or when visual evidence is limited. In addition, when parts of the body are occluded by external objects (\eg tables or obstacles in the scene), the estimated motion can become incomplete or inaccurate due to missing visual cues. We also observe that when the camera captures the scene from a rear viewpoint, motion estimation becomes less reliable due to limited visibility and frequent self-occlusion. 
These issues further reduce the usability of the extracted motion for downstream control. A representative example of these failure cases is shown in Fig.~\ref{fig:motion_failure}.

Furthermore, wrist motion estimation still shows notable failure cases. In scenarios where the hand should perform a vertical grasp (\eg grasping a bottle), the generated video depicts correct interactions, but the estimated motion often results in a horizontally oriented wrist. This suggests the model struggles to infer fine-grained wrist rotations, especially when visual cues are ambiguous or weak in monocular views. Such errors can propagate to downstream control, leading to misaligned grasps and reduced manipulation success.

\paragraph{Failure of Whole-body Execution}
At the execution stage, errors primarily arise from mismatches between estimated motion and the robot’s physical constraints. Common issues include inaccurate hand height during interactions and deviations in movement distance during navigation. These discrepancies can affect task completion, especially for tasks requiring precise spatial alignment or contact. In multiple cases, we observe that the robot fails to reach or properly interact with target objects due to insufficient height alignment; as mentioned in Section~\ref{sec:difficult_level}, we place small supporting bases underneath the target objects to slightly elevate them, as a current compromise.

\subsubsection{Ablation Case Studies}
\label{sec:ablation}
\paragraph{Video Generation Models}
We evaluate several off-the-shelf video generation models, including Veo 3.1~\citep{wiedemer2025video}, Kling 3~\citep{team2025kling0omni}, and Wan 2.6~\citep{wan2025wan0}, within the implemented \projname{} system pipeline. Although all models are capable of producing task-relevant video sequences, we observe noticeable differences in task completion, physical simulation fidelity, motion stability, and adherence to the input prompts. In particular, Kling 3 consistently generates more coherent and physically plausible motion, with fewer visual artifacts and better alignment with the intended progression of future action chains. In contrast, Veo 3.1 and Wan 2.6 more frequently exhibit issues such as motion drift, inconsistent human-object interactions, spurious erroneous generation, or degraded end-state stability. Based on these empirical observations, we adopt Kling 3 as the primary video generation model in our final system.

\paragraph{Retargeting Ablation}
We study the effect of introducing an intermediate motion retargeting stage between the estimated human motion and real robot execution. Given reconstructed SMPLX trajectories, retargeting methods such as GMR~\citep{araujo2025retargetingmattersgeneralmotion} and OmniRetarget~\citep{yang2025omniretargetinteractionpreservingdatageneration} aim to map human kinematics to robot DoFs, while improving temporal smoothness. In our experiments, applying retargeting does frequently lead to more fluent whole-body motion and reduces high-frequency jitter, however, also commonly introducing noticeable spatial deviations.

These deviations arise from multiple factors. First, the global position of the estimated human motion is inherently noisy, often exhibiting drift and foot sliding; retargeting tends to attempt to correct these artifacts, thus changing the overall motion trajectory. Second, the embodiment mismatch between humans and robots (\eg differences in body scale and limb proportions) leads to differences in step length and motion geometry after retargeting.

In contrast, directly feeding the estimated motion into the controller preserves higher geometric fidelity and improves positional accuracy, despite introducing small discontinuities from reconstruction noise. Since both navigation and manipulation tasks in our setting require accurate spatial alignment with target locations, we prioritize precision over smoothness and therefore do not adopt retargeting in our implemented system. Moreover, the utilized SONIC backend is able to handle such noise without the robot motion failing catastrophically.

\paragraph{Third-person Perspective Ablation}
We study the impact of camera viewpoint on the success rate of our pipeline. We observe that different perspectives are better suited for different types of tasks. For navigation tasks, exocentric views from back-to-front tend to yield higher success rates, as they provide clearer aligned information about the robot’s movement direction and spatial layout. In contrast, for manipulation tasks, front-facing views are more effective, as they better capture hand motions and object interactions. These results highlight the importance of viewpoint selection in video-driven humanoid control, suggesting that task-aware perspective design can improve overall system performance.

\paragraph{Motion Estimation Ablation}
We compare two motion estimation methods in our pipeline: CRISP~\cite{wang2026crispcontactguidedreal2simmonocular} and GENMO~\citep{li2025genmo0}. CRISP jointly reconstructs human motion and scene geometry by enforcing consistency between 2D observations and 3D physical priors. Empirically, GENMO produces motion trajectories that are comparable in quality to CRISP, while being significantly more efficient and stable in practice. Given this favorable trade-off, we adopt GENMO as the default motion estimation method in our system.

\subsubsection{Latency Analysis}

Our current framework follows an offline pipeline, where each stage must be completed before execution can begin. Specifically, given an initial observation, the system sequentially performs image style transfer, third-person video generation, and motion estimation, after which the recovered motion is executed by the humanoid robot. As a result, the overall latency is dominated by the cumulative time of these upstream processing steps.

\begin{wraptable}{r}{0.5\textwidth}
\vspace{-0.5em}
\centering
\resizebox{0.5\textwidth}{!}{%
\begin{tabular}{lcc}
\toprule
\textbf{Module} & \textbf{Metric} & \textbf{Avg. Time (s)} \\
\midrule
Robot-to-Human Embodiment Transfer & per request &  10.7\\
Task-to-Action Decomposition and Prompt Construction & per request &  2.5\\
\midrule
Task- and Environment-Generalizable Video Generation & per sec (video) &  13.2\\
Whole-body Motion Estimation & per sec (video) &  2.9\\
Hand Motion Estimation & per sec (video) &  16.4\\
\bottomrule
\end{tabular}
}
\caption{Average runtime of different components in the \projname{} pipeline. We report per-request latency for single-shot modules, and per-second processing time for video-dependent modules.}
\label{tab:latency}
\vspace{-1em}
\end{wraptable}

To better understand the computational cost of the pipeline, we report the average runtime of each component, including (1) robot-to-human embodiment transfer, (2) video generation, and (3) motion estimation. The results are summarized in Table~\ref{tab:latency}. This analysis highlights that video generation constitutes a critical bottleneck in the pipeline, followed by motion estimation, while the embodiment transfer step incurs relatively minor overhead. These findings motivate future work on improving the efficiency of generative models and exploring more streamlined pipelines for real-time or streaming execution.

\section{Discussion}
\label{sec:discussion}

Here we discuss several important directions for future research, many of which also reflect current limitations of our system, to further advance the proposed paradigm and encourage broader exploration from the community. While \projname{} demonstrates the feasibility of leveraging exocentric video generation for modeling and executing interaction-rich humanoid behaviors, it also exposes challenges across video generation quality, motion estimation accuracy, and embodied control robustness, which currently constrain overall system performance. Addressing these limitations will require improvements in physical realism, video-to-motion translation, and adaptive control. We outline several directions that we believe are critical for scaling this paradigm towards more robust and generalizable humanoid systems.

\subsection{Closed-loop and Scene-aware Whole-body Control with Exocentric Motion Reference}

In this work, the generated exocentric videos primarily provide open-loop motion references for downstream execution, where the robot replays trajectories without explicit awareness of the surrounding environment. While this approach may be sufficient for controlled settings, this design limits robustness under perception noise, localization errors, dynamic obstacles, and other real-world uncertainties. A key future direction is to develop closed-loop, scene-aware whole-body control that integrates perception with control.

Instead of treating the estimated motion as a fixed trajectory, the controller could use it as a high-level reference while continuously adapting behavior based on online visual observations and proprioceptive feedback. This would allow the robot to preserve task intent and interaction structure encoded in the generated videos, while dynamically adjusting foot placement, body posture, hand motion, and timing according to the environment. Such a formulation bridges generative task imagination with feedback-driven control, enabling more robust execution of long-horizon, contact-rich, and interaction-intensive humanoid behaviors.

\subsection{Physically Realistic Video Generation}

The effectiveness of our pipeline fundamentally depends on the realism of generated videos. Although existing video generation models can capture high-level semantic information reasonably well, their output often emphasizes visual plausibility rather than being physically realizable. This limitation is particularly critical for downstream motion estimation and execution, as spatiotemporal inconsistencies, incorrect contact relationships, or violations of kinematic constraints in the generated videos can directly degrade the accuracy of trajectory recovery and reduce stability of real-world execution. In this sense, insufficient physical realism is not merely a generation-quality issue, but a systemic bottleneck that limits the overall performance of the framework.

Future work focus on incorporating stronger physical (and other causality) priors into video generation models, including object permanence, contact dynamics, geometric consistency, and motion constraints. In addition, evaluation metrics should be better aligned with embodied tasks, assessing not only visual fidelity but also physical plausibility and executability. Improving these aspects is essential for enabling more reliable transfer from generated videos to executable humanoid behaviors.

\subsection{Streaming and Real-time Task Imagination and Execution}

Our current framework follows an offline pipeline, where the system generates a complete third-person video, estimates motion, and then executes it on the humanoid robot. While effective for validation, this design limits adaptability to dynamic environments and execution errors. A promising direction is to develop a streaming formulation in which task imagination and execution proceed jointly. Instead of generating the full trajectory upfront, the system could produce short-horizon video segments conditioned on current observations and task progress, and immediately convert them into executable motion. Such a pipeline would enable continuous re-planning, allowing the robot to adapt to environmental changes and recover from deviations. Achieving this requires efficient video generation, low-latency motion estimation, and real-time integration with whole-body controllers, but could significantly improve the practicality of video-driven humanoid systems.

\subsection{From First-person to Third-person Generation}

Our current formulation assumes access to third-person observations that capture the full-body motion of the humanoid, the surrounding environment, and task-relevant objects. In many practical settings, such as home environments, this assumption can be partially satisfied by deploying external monitoring cameras that provide exocentric views of the robot and target objects. However, such infrastructure may not always be available or reliable in more general scenarios.

An important future direction is therefore to develop models that can generate third-person task execution videos from first-person observations, enabling scalable deployment using onboard sensors and proprioception alone. By reconstructing or imagining exocentric views from egocentric perception, the system could retain the advantages of third-person interaction modeling while relaxing the requirement for external cameras. This would significantly broaden the applicability of the framework and make it more practical for real-world deployment.

\subsection{Improved Video-to-Motion Translation}

A critical component of the pipeline is the conversion from generated videos to executable motion. Compared with the video generation stage, motion estimation introduces a major source of uncertainty and errors. Existing approaches typically decompose the process into multiple stages, such as pose estimation, tracking, trajectory fitting, and motion retargeting. While perhaps convenient, such modular designs are prone to error accumulation, especially under occlusion, viewpoint changes, rapid motion, or temporal artifacts in generated videos. As a result, intermediate representations can become noisy or ambiguous, degrading the accuracy and executability of the recovered motion. For embodied agents, this issue is even more severe, as visually plausible motions may still violate kinematic or dynamic constraints.

Future work must explore more accurate and adaptive video-to-motion translation models that can explicitly account for embodiment alignment and control feasibility. A promising direction is to develop learning-based approaches that jointly optimize motion recovery and downstream control objectives, ensuring that inferred motions are both visually consistent and physically executable. Incorporating embodiment-aware priors and task-oriented training signals may further improve robustness and generalization across diverse settings.

Meanwhile, accurate wrist motion extraction is crucial for manipulation, yet remains challenging due to ambiguity in fine-grained transformations, especially under occlusion or monocular viewpoints. As observed, current models often fail to recover correct wrist orientations (\eg vertical grasps predicted as horizontal), indicating that such rotations are weakly constrained by visual cues. Improving robustness may require geometric or kinematic priors, temporal consistency, and object-aware constraints to better align hand orientation with interaction goals. Addressing these issues is essential for bridging the gap between visually plausible motion and reliable control.

\subsection{Robot-centric Video Generation}

In \projname{}'s current implementation, the system relies on style transfer to map robot appearances into a human-like domain before video generation. While effective in practice, this step introduces an additional domain change that may weaken the consistency between generated motions and executable robot behaviors. An important future direction would be to develop more general video generation models capable of directly operating on different embodiments (like robots), eliminating the need for human-centric intermediate representations.

Importantly, such robot-capable (or robot-centric) generation should not simply render robots as visual subjects or inherit rigid, unnatural motion patterns from existing robot data. Instead, models should generate motions that are structurally compatible with robotic embodiments while remaining smooth, continuous, and physically coherent. These properties should be reflected in temporal consistency, multi-view stability, and physically plausible interactions.
More broadly, robot-centric video generation offers a pathway toward more faithful modeling of the relationship between embodiment morphology, environmental interaction, and action outcomes.

\subsection{Benchmarking Video-driven Humanoid Control}

A key limitation in this emerging paradigm is the lack of standardized benchmarks for evaluating video-driven humanoid control systems. Current evaluations are often fragmented, focusing separately on video generation quality, motion reconstruction accuracy, or control performance, making it difficult to assess end-to-end effectiveness. We argue for the development of unified benchmarks that jointly evaluate the full pipeline, including video generation, motion estimation, and embodied execution. Such benchmarks should measure not only visual realism, but also physical plausibility, interaction correctness, and task success in real-world settings. Establishing standardized evaluation protocols and datasets would provide a common ground for comparing methods, reveal system-level trade-offs, and accelerate progress toward more reliable and generalizable humanoid control.

\section{Related Work}

\subsection{World Action Models for Robotics}

World action models~\citep{yuan2026fastwam,parker2024genie,bruce2024genie,assran2023self} aim to capture future observations under agent actions, thereby furnishing predictive structure for control, policy learning, and decision making. Following early action-conditioned dynamics models in robotics~\citep{wu2023daydreamer}, recent research has increasingly moved toward visually grounded predictive modeling to better represent fine-grained interaction dynamics and scene-dependent behavior. A prominent direction uses future visual prediction as an intermediate substrate for policy learning, leveraging predicted or generated videos as behavioral priors, synthetic supervision, or joint action-visual training signals~\citep{liang2025video,mei2026video,bharadhwaj2025gen2act,gu2026say,li2026causal,cen2025worldvla,cen2025rynnvla,ye2026world,jang2025dreamgen}. Complementary efforts focus on short-horizon predictive context, replacing raw-frame prediction with geometric, semantic, or latent future representations to balance foresight quality and computational efficiency, and integrating such predictive objectives into VLA pretraining and online action refinement~\citep{zhanggenerative,huang2025ladi,hu2025video,sun2026vla,liu2026self,yang2026chain}. 

Despite this progress, current methods still struggle to jointly achieve long-horizon prediction, efficient inference, and physically precise anticipation, which motivates more practical world action models for humanoid robotic control.
\subsection{Physics Simulation based on World Models}
Learned world-model-based simulation has emerged as a complementary alternative to traditional physics simulators for robot learning. Whereas classical simulators provide controllable and interpretable environments under simplified assumptions about geometry, contact, and material dynamics, learned simulators model environment evolution directly from data and therefore offer a flexible substrate for planning, optimization, and policy adaptation in visually rich settings. Recent work has increasingly treated world models as trainable interactive simulators for downstream policy learning, using them to provide embodied supervision, convert generated visual plans into dynamically feasible trajectories, and support post-training improvement of vision-language-action policies~\citep{lin2025physbrain,ziakas2026grounding,menggrounding,liu2026world}. This line has further expanded toward longer-horizon interaction by strengthening memory mechanisms, improving rollout stability, and enhancing the reliability of policy adaptation~\citep{team2026advancing,yu2026mosaicmemhybridspatialmemory,jiang2026wovr,gigabrainteam2026gigabrain05mvlalearnsworld,torne2026memmultiscaleembodiedmemory}. In parallel, learned simulation is also being explored as a scalable tool for synthetic data generation and embodiment transfer, where transferable robot experience is constructed from cross-platform robot videos, geometry-consistent visual transfer, and transformed human demonstrations~\citep{bai2025roboswap,liu2025robotransfer,lepertmasquerade}. However, recent benchmark studies indicate that current methods remain limited in action-conditioned causality, physical plausibility, and temporal coherence, revealing a persistent gap between visually plausible generation and reliable embodied simulation~\citep{guo2025r,deng2026rethinking}.

In contrast to these approaches that focus on learning simulators for predictive modeling or policy training, our work leverages generative video models as an intermediate interface for directly synthesizing interaction dynamics and translating them into executable humanoid behaviors, without requiring an explicit world model or simulation loop.
%

\subsection{Generalizable Humanoid Control}

Generalizable humanoid control aims to endow humanoid robots with robust whole-body behaviors across tasks, environments, and embodiments under perception noise, motion variation, and distribution shift. Compared with fixed locomotion or task-specific manipulation, this problem is substantially more challenging because humanoids must jointly coordinate locomotion, balance, upper-body motion, active perception, and task interaction within a unified control framework. Recent research mainly progresses along two directions. One direction focuses on grounding high-level semantics into executable whole-body behavior, for example by learning instruction-conditioned vision-language-action policies, latent loco-manipulation controllers, or spatially aware egocentric action representations for long-horizon execution~\citep{bai2026egoactor0,li2026fromw1generalhumanoidwholebody,jiang2025wholebodyvlaunifiedlatentvla}. Closely related work improves generalization through scalable human-centric supervision, transferring egocentric demonstrations, monocular human videos, motion capture, and transformed third-person videos into humanoid skills without relying solely on expensive robot teleoperation or task-specific reward engineering~\citep{shi2026egohumanoidunlockinginthewildlocomanipulation,allshire2025visualimitationenablescontextual,yang2026zerowbclearningnaturalvisuomotor,yang2025x}. The other direction emphasizes controller generalization itself, seeking whole-body policies that remain stable and adaptable under motion diversity and embodiment variation through residual adaptation, morphology-aware policy design, expert-to-generalist distillation, and unified generative pretraining for dynamic motion tracking~\citep{pertsch2025fastefficientactiontokenization,xue2026scalablegeneralwholebodycontrol,wang2025expertsgeneralistgeneralwholebody,wang2026omnixtremebreakinggeneralitybarrier, luo2025sonicsupersizingmotiontracking}. 

Overall, existing studies have substantially advanced semantic grounding, human-data transfer, and controller generalization, yet major challenges remain in long-horizon compositionality, cross-embodiment robustness, and physically consistent real-world deployment.

\section{Conclusion}

We present \projname{}, an end-to-end framework and system that leverages exocentric video generation as an interface for modeling interaction-rich humanoid behaviors. Instead of directly mapping task instructions to low-level robot actions, \projname{} first synthesizes third-person task execution videos that capture coordinated interactions between the humanoid, its environment, and task-relevant objects, and then converts these generated videos into executable behaviors through motion estimation and general motion tracking. Our real-world experiments demonstrate the feasibility of using generated videos as actionable intermediate representations for humanoid control without task-specific data collection.

More broadly, \projname{} suggests a promising direction for connecting generative video models with embodied humanoid control. While current limitations remain in physical realism, video-to-motion translation, open-loop execution, and manipulation-intensive tasks, we believe this work provides an initial step toward scalable humanoid systems that use generative models to imagine, structure, and execute complex interactions in the physical world.

\bibliography{iclr2026_conference}
\bibliographystyle{iclr2026_conference}

\newpage
\appendix
\section{Appendix}

\subsection{Action Decomposition Organization}
\label{app:action_decomposition}
In this section, we present the prompt used for action decomposition in Fig.~\ref{fig:action_decomposition_prompt}. The prompt converts high-level task goals into temporally ordered, robot-style action chain paragraphs, explicitly specifying intermediate motions and interaction details. This structured representation provides a consistent and visually grounded interface for third-person video generation, ensuring that the synthesized actions remain physically plausible and aligned with downstream motion extraction and execution.

\subsection{Embodiment Transfer Prompt}
\label{app:robot_to_human_prompt}
In this section, we present the system prompt used for robot-to-human embodiment transfer in Fig.~\ref{fig:robot_to_human_replacement_prompt}. The prompt enforces strict constraints on pose, orientation, and scale to ensure one-to-one alignment between the robot and the generated human subject, while preserving the original scene and camera configuration. The initial-state information is produced by Gemini 3.1 according to the prompt, which requests structured initial-state information extracted from the first frame, including the target person’s position, body pose, and camera-relative orientation, as well as the positions and orientations of salient objects, and the initial spatial relationships among the target person, salient objects, and the surrounding environment. This information is subsequently incorporated into the embodiment transfer prompt as important guidance. This design allows the transformed image to remain compatible with downstream video generation and motion extraction, both of which rely on human-centric visual priors.

\begin{figure*}
    \centering
    \setlength{\textfloatsep}{4pt}
    \begin{systemprompt}
    \footnotesize
    \setlength{\parskip}{0pt}
    \setlength{\parindent}{0pt}

    \textbf{Action Decomposition Prompt: Task Goal to Robot-Style Action Chain}\\[-1pt]
    \textit{You are a professional robotic action planner}. Convert each high-level user task goal into a concise, temporally ordered, robot-style action chain paragraph for third-person action video generation and robot motion retargeting.\\[-1pt]

    \textbf{Input.}\\[-1pt]
    One or more task goals, optionally with scene information, target objects, initial state, or desired final state.\\[-1pt]

    \textbf{Decomposition Requirements.}
    \begin{itemize}[leftmargin=*, topsep=0pt, itemsep=0pt, parsep=0pt, partopsep=0pt]
        \item Decompose each task into physically plausible sub-actions in chronological order.
        \item Describe robot-centric motion: torso/base orientation, arm movement, end-effector alignment, gripper state, contact, object motion, release, and final stabilization.
        \item Make implicit intermediate motions explicit, especially reaching, grasping, lifting, moving, placing, opening, closing, and retracting.
        \item Keep each action visually concrete and suitable for third-person video generation.
        \item Do not introduce new objects, tools, scene elements, or task goals beyond the input.
        \item Avoid emotional, cinematic, or decorative language; focus on functional robot motion.
    \end{itemize}

    \textbf{Output Format.}\\[-1pt]
    For each task goal, output exactly one paragraph. Do not use bullet points, numbered steps, JSON, explanations, or extra commentary. Each paragraph should contain 4--8 ordered sub-actions using transition words such as ``first'', ``then'', ``next'', and ``finally''.\\[-1pt]

    \textbf{TASK.}\\[-1pt]
    Convert each provided task goal into one robot-style decomposed action chain paragraph following the requirements above.

    \end{systemprompt}
    \caption{The prompt used to decompose high-level task goals into robot-style action chain paragraphs.}
    \label{fig:action_decomposition_prompt}
\end{figure*}

\begin{figure*}
    \centering
    \setlength{\textfloatsep}{2pt}
    \begin{systemprompt}
    \scriptsize
    \setlength{\parskip}{0pt}
    \setlength{\parindent}{0pt}

    \textbf{Image Editing Prompt: Humanoid Robot to Human}\\[-2pt]
    \textit{You are a professional image editing system}. Your task is to replace the bipedal robot in the input image with a single male human subject, while preserving the scene exactly and maintaining precise pose, orientation, and scale alignment.\\[-2pt]

    \textbf{Style.}\\[-2pt]
    High-definition, photorealistic, seamless integration.\\[-2pt]

    \textbf{Initial State Information.} \\[-2pt]
    [INITIAL STATE INFORMATION]: The structured initial-state information extracted from the first frame, including the target person's position, body pose, camera-relative orientation, the positions and orientations of salient objects, and the initial spatial relations among the target person, salient objects, and the surrounding environment. \\

    \textbf{Mandatory Alignment Principle.}\\
    Execute an exact, 1:1, degree-perfect replication of the robot's full-body spatial pose, orientation, and kinematic configuration as seen in the input image. This includes the head, torso, arms, legs, and feet angles. The highest-priority rule is front/back consistency: if the robot is front-facing, the human subject must be front-facing; if the robot is back-facing, the human subject must be back-facing. Scale must match exactly. In particular, the human subject's shoulder line and waistline must align precisely with the spatial coordinates, height, and position of the robot's original shoulder line and waistline.\\[-2pt]

    \textbf{Subject Details.}
    \begin{itemize}[leftmargin=*, topsep=0pt, itemsep=-1pt, parsep=0pt, partopsep=0pt]
        \item A single bipedal male human subject.
        \item The subject should preserve the robot's distinctive proportions while incorporating a Hobbit-like body profile.
        \item The shoulders should remain notably broad.
        \item The limbs should appear elongated but balanced.
        \item The torso must be significantly shorter than that of a normal human, creating a compact, short-waisted, Hobbit-like silhouette.
        \item The waistline should be extremely slender and well-defined.
        \item The overall physique should be compact, streamlined, and non-muscular.
        \item The overall height should match the robot's apparent height, which is roughly about one head shorter than a normal adult human.
    \end{itemize}

    \textbf{Attire.}
    \begin{itemize}[leftmargin=*, topsep=0pt, itemsep=-1pt, parsep=0pt, partopsep=0pt]
        \item The subject should wear an ultra-fitted technical sports compression base layer.
        \item The garment should appear as a single-piece outfit with minimal seams.
        \item Use plain, solid dark tones, such as charcoal grey or matte black.
        \item The outfit must contain no brand names, text, logos, graphics, identifier tags, or chest labels.
        \item Loose fabrics are strictly prohibited, including cloaks, capes, robes, overcoats, or any clothing that obscures the subject's physique, proportions, or the background.
    \end{itemize}

    \textbf{Hands.}
    \begin{itemize}[leftmargin=*, topsep=0pt, itemsep=-1pt, parsep=0pt, partopsep=0pt]
        \item Both hands must be empty.
        \item The hands should be in a natural, open, relaxed pose.
        \item The hand state should mimic the empty state of the robot's grippers.
        \item The subject must not hold any external object, including tools, cables, cups, or any other item.
    \end{itemize}

    \textbf{Important Constraints.}
    \begin{itemize}[leftmargin=*, topsep=0pt, itemsep=-1pt, parsep=0pt, partopsep=0pt]
        \item Preserve the original scene with absolutely no changes.
        \item Maintain exact pose, orientation, body placement, and scale consistency with the robot.
        \item Ensure the generated human remains consistent with the robot's unusual body proportions and shorter-than-normal height.
        \item Do not introduce any additional objects, accessories, or environmental modifications.
    \end{itemize}

    \textbf{TASK.}\\
    Replace the robot in the input image with the described male human subject, following all constraints above, while preserving the scene exactly.

    \end{systemprompt}
    \caption{The system prompt for robot-to-human replacement under strict pose, orientation, and scene-preservation constraints.}
    \label{fig:robot_to_human_replacement_prompt}
\end{figure*}

\subsection{Action-Chain-Conditioned Generation Prompt Construction}
\label{app:fusion_prompt}

In the Fig.~\ref{fig:multimodal_action_description_prompt}, Given the decomposed action chain, we construct the final video generation prompt by grounding the planned action sequence in the initial observation and the original task goal. Specifically, the action chain provides the temporal structure of the intended behavior, while the initial-state information specifies the target person's position, body pose, camera-relative orientation, salient object locations, and their spatial relations. We combine these inputs to produce a spatial- and task-aware action description, which is then inserted into a fixed generation prompt template. The template further constrains camera viewpoint, scene consistency, motion stability, execution style, and final task completion, improving temporal coherence and reducing unintended visual changes.
\begin{figure*}[t]
    \centering
    \setlength{\textfloatsep}{4pt}
    \begin{systemprompt}
    \footnotesize
    \setlength{\parskip}{0pt}
    \setlength{\parindent}{0pt}

    \textbf{Multimodal Action Description Construction Prompt}\\[-1pt]

    You are a professional video-generation prompt engineer. Given \texttt{[VISUAL OBSERVATION]}, \texttt{[ACTION CHAIN]}, and \texttt{[ACTION GOAL]}, construct \texttt{[SCENE- AND TASK-AWARE ACTION DESCRIPTION]} for third-person action video generation.\\[-1pt]

    \textbf{Input.}\\[-1pt]
    \texttt{[VISUAL OBSERVATION]}: The initial frame or scene description, including scene layout, visible objects, target subject, object locations, and initial spatial relations.\\[-1pt]

    \texttt{[ACTION CHAIN]}: A temporally ordered sequence of intended sub-actions.\\[-1pt]

    \texttt{[ACTION GOAL]}: The high-level task objective to be completed.\\[-1pt]

    \textbf{Construction Requirements.}\\[-1pt]
    \textbullet\ Preserve \texttt{[ACTION GOAL]} and the temporal order of \texttt{[ACTION CHAIN]}.\\[-1pt]
    \textbullet\ Ground every action in \texttt{[VISUAL OBSERVATION]}, including object locations, reachable targets, subject position, and spatial constraints.\\[-1pt]
    \textbullet\ Make the action progression explicit, continuous, and physically plausible.\\[-1pt]
    \textbullet\ Describe human-object interactions only when the relevant objects are visible or implied by \texttt{[VISUAL OBSERVATION]}.\\[-1pt]
    \textbullet\ Do not introduce new objects, scene elements, tools, or goals beyond \texttt{[VISUAL OBSERVATION]} and \texttt{[ACTION GOAL]}.\\[-1pt]
    \textbullet\ Ensure the final state clearly satisfies \texttt{[ACTION GOAL]}.\\[-1pt]
    \textbullet\ Write the result as a concise action description suitable for insertion into the Action section of a video generation prompt.\\[-1pt]

    \textbf{Output.}\\[-1pt]
    \texttt{[SCENE- AND TASK-AWARE ACTION DESCRIPTION]}: A single concise action description grounded in the visual scene and aligned with the task goal. 

    \end{systemprompt}
    \caption{The prompt used to construct a scene- and task-aware action description from multimodal inputs.}
    \label{fig:multimodal_action_description_prompt}
\end{figure*}

\subsection{B-Level Navigation Task Generation Prompt}
\label{app:b_level_navigation_prompt}
In this section, we present the prompt template used for B-level navigation task generation (Fig.~\ref{fig:b_level_navigation_prompt}). The prompt enforces a fixed-camera, scene-consistent setting and specifies natural, physically plausible navigation behaviors. This design ensures that the generated videos remain stable, temporally coherent, and suitable for downstream motion extraction and execution.

\begin{figure*}[t]
    \centering
    \setlength{\textfloatsep}{4pt}
    \begin{systemprompt}
    \footnotesize
    \setlength{\parskip}{0pt}
    \setlength{\parindent}{0pt}

    \textbf{(B-Level Navigation Task Generation Prompt)}\\[-1pt]
    \textit{Generate a task-consistent human action video under a fixed third-person camera view.}\\[-1pt]

    \textbf{Shot.}\\[-1pt]
    A 10-second locked-off static shot of a real indoor scene. The camera remains completely fixed throughout the video, with no movement, zoom, reframing, or perspective change.\\[-1pt]

    \textbf{Scene.}\\[-1pt]
    The environment shown in the first frame remains unchanged. All furniture, objects, and background elements stay fixed unless explicitly involved in the target task.\\[-1pt]

    \textbf{Subject.}\\[-1pt]
    The target person is a man in a skintight outfit, visible in the first frame at the initial position.\\[-1pt]

    \textbf{Action.}\\[-1pt]
    This is a navigation task. Starting from the exact first-frame position, the target person performs the following action: \texttt{[NAVIGATION ACTION]}.\\[-1pt]

    \textbf{Motion.}\\[-1pt]
    The motion should be natural, continuous, stable, and physically realistic. Maintain realistic balance and ground contact, with no foot sliding, jitter, abrupt motion, exaggerated gestures, or unnecessary movement. If walking is involved, use a smooth, moderate gait.\\[-1pt]

    \textbf{Consistency.}\\[-1pt]
    Maintain first-frame scene geometry and fixed-camera consistency. Do not change the scene layout, object placement, background, viewpoint, framing, scale, or perspective.\\[-1pt]

    \textbf{End State.}\\[-1pt]
    After completing the navigation task, the target person reaches the specified destination and stands in a stable final posture. The final frame should show the unchanged environment and the successfully completed navigation outcome.\\[-1pt]

    \textbf{TASK.}\\[-1pt]
    Generate a 10-second static-view action video following the requirements above, where \texttt{[NAVIGATION ACTION]} is instantiated with the user-provided navigation goal.

    \end{systemprompt}
    \caption{The prompt template used for B-level navigation task generation.}
    \label{fig:b_level_navigation_prompt}
\end{figure*}

\subsection{A/S-Level Advanced Task Generation Prompt}
\label{app:as_level_task_prompt}
In this section, we present the prompt template used for A/S-level advanced task generation (Fig.~\ref{fig:as_level_task_prompt}). Compared to B-level navigation, this prompt incorporates additional constraints for manipulation and long-horizon execution, explicitly specifying task structure, interaction details, and robot-like motion characteristics. This formulation encourages physically plausible, temporally coherent behaviors that are suitable for downstream motion extraction and execution.

\begin{figure*}[t]
    \centering
    \setlength{\textfloatsep}{4pt}
    \begin{systemprompt}
    \footnotesize
    \setlength{\parskip}{0pt}
    \setlength{\parindent}{0pt}

    \textbf{(A/S-Level Advanced Task Generation Prompt)}\\[-1pt]
    \textit{Generate a task-consistent human action video under a fixed third-person camera view for advanced manipulation or long-horizon tasks.}\\[-1pt]

    \textbf{Shot.}\\[-1pt]
    A 10-second locked-off static shot of a real indoor scene. Fixed viewpoint, framing, focal length, zoom, and perspective. No camera movement or reframing.\\[-1pt]

    \textbf{Scene.}\\[-1pt]
    The environment shown in the first frame remains completely unchanged. All objects, furniture, and background elements stay fixed. Any non-target people remain perfectly still.\\[-1pt]

    \textbf{Subject.}\\[-1pt]
    The target person is a man in a skintight outfit, visible in the first frame at the initial position.\\[-1pt]

    \textbf{Task.}\\[-1pt]
    From the exact first-frame position, the target person performs the following task: \texttt{[TASK ACTION CHAIN]}.\\[-1pt]

    \textbf{Motion.}\\[-1pt]
    Motion should be natural, continuous, stable, and physically realistic. Maintain realistic balance and ground contact, with no foot sliding, unnecessary gestures, abrupt motion, exaggerated movement, or unintended contact. If walking is involved, use a smooth, moderate gait.\\[-1pt]

    \textbf{Execution.}\\[-1pt]
    Follow the task-specific execution block: \texttt{[TASK-SPECIFIC EXECUTION BLOCK]}. The action should be performed in a humanoid-robot-like style: short, cautious, smooth, and deliberate movements; clear alignment before interaction; controlled reaching; precise contact and grasp behavior; stable object transfer when manipulation is involved; and a clean return to a neutral final posture.\\[-1pt]

    \textbf{Consistency.}\\[-1pt]
    Maintain first-frame scene geometry and fixed-camera consistency. Do not introduce scale drift, perspective warping, unintended object motion, extra human motion, or background changes.\\[-1pt]

    \textbf{End State.}\\[-1pt]
    After completing the task, the target person comes to a stable final posture. The final frame shows the unchanged environment and the successfully completed task outcome.\\[-1pt]

    \textbf{TASK.}\\[-1pt]
    Generate a 10-second static-view action video following the requirements above, where \texttt{[TASK ACTION CHAIN]}, and \texttt{[TASK-SPECIFIC EXECUTION BLOCK]} are instantiated with the user-provided advanced task specification.

    \end{systemprompt}
    \caption{The prompt template used for A/S-level advanced task generation.}
    \label{fig:as_level_task_prompt}
\end{figure*}
\end{document}